%% file: main.tex
\documentclass{article}
\usepackage[margin=1in]{geometry}
\usepackage{graphicx} %
\usepackage{xcolor}
\usepackage{xspace}
\usepackage{appendix}
\usepackage{booktabs}
\usepackage[numbers]{natbib}
\usepackage{amsmath}
\usepackage{microtype}
\usepackage[hidelinks]{hyperref}
\usepackage{cleveref}
\crefalias{section}{appendix}

\newcommand{\ourbench}[1]{\textsc{NewsBench}\xspace}

\input{generated/paper_metrics/latest/metrics_macros}
\title{\ourbench{}: Expert-Grounded Evaluation of Epistemic Quality in AI News Reporting}

\begin{document}

\begin{center}
{\LARGE\bfseries \ourbench{}: Expert-Grounded Evaluation of Epistemic Quality in AI News Reporting
 \par}
\vspace{12pt}
{\normalsize
Matt Wilde,$^{1}$
Jillian Fisher,$^{1,3}$
Andrew B.\ Hall,$^{1,2}$
Kathryn Salam,$^{1}$
Robbie Goldfarb$^{1}$
\par}
\vspace{6pt}
{\small\itshape
$^{1}$Forum AI \quad
$^{2}$Stanford University
$^{3}$University of Washington
\par}
\vspace{16pt}
May 2026
\end{center}

\begin{abstract}
    As frontier AI systems increasingly serve as sources of political and current-events information, evaluating the epistemic quality of their responses is now especially consequential. Existing evaluations often focus narrowly on ideological alignment, rely on measures not validated for this setting, or depend on reference labels derived from public opinion or non-expert annotation, potentially producing misleading or miscalibrated results. To address these limitations, we introduce \ourbench{}, an expert-grounded benchmark that reframes evaluation around editorial standards developed by senior practitioners in journalism, policy, and intelligence analysis. \ourbench{} operationalizes quality across three dimensions, source quality, factuality, and neutrality, using a two-phase pipeline in which domain experts first define and calibrate evaluation guidelines, which are then transferred to automated AI judges validated against expert-adjudicated gold labels. The benchmark includes approximately $3,000$ evaluation prompts, with roughly $500$ refreshed monthly to capture emerging current events, roughly $2,500$ human expert-annotated items for calibration and validation, and a suite of AI judges designed to enable scalable evaluation. To validate our methodology, we report both inter-rater reliability among human experts and AI judge–human agreement across all three dimensions. Finally, we apply \ourbench{} to four frontier models, revealing distinct and non-overlapping failure profiles in sourcing, factuality, and neutrality.
\end{abstract}

\input{sections/1_introduction}

\input{sections/2_method}
\input{sections/3_results_validation}
\input{sections/4_results_AIpreformance}

\bibliographystyle{unsrtnat}
\bibliography{main} %

\section*{AI Acknowledgment}

During the preparation of this work, the authors used Claude Opus 4.7 to help summarize material and improve the readability of the manuscript. After using this tool, the authors reviewed and edited the content as needed and take full responsibility for the content of the publication.

\end{document}

%% file: sections/1_introduction.tex
\section{Introduction}

Frontier AI models are increasingly capable across a wide range of tasks, but growing evidence shows they can exhibit political bias \cite{Li2023TheDS, forbes_bias_llm, westwood2025slant, fulay-etal-2024-relationship}, and this bias can influence users' behaviors \cite{fisher-etal-2025-biased, potter2024hidden, nature_persuasion}.
These risks are particularly acute in \textbf{news and current-events settings}, where users rely on model outputs to form beliefs about the world. As a result, it is critical to develop evaluations that assess not only whether models avoid partisan bias, but whether they produce high-quality, factual, and appropriately framed news responses. Such evaluations must be reliable, generalizable, and responsive to a rapidly changing information environment. Yet constructing them is difficult, and the stakes of getting it wrong are high. Moreover, there is a credibility problem, as the organizations building these systems are not well-positioned to evaluate their own outputs \cite{fisher2025neutrality}, and governments are increasingly demanding independent evidence that these systems meet appropriate standards \cite{s2938_119th, nist_caisi}. In response, Forum AI aims to construct independent, expert-guided evaluations of AI systems for societally consequential domains, beginning with news and political information.

Despite growing attention to politically sensitive content in AI systems, \textbf{existing evaluations are fundamentally limited because they are not grounded in expert judgment} and therefore lack the foundations needed to assess news response quality in a principled way. As a result, they struggle to define what should be measured, how it should be measured, and what constitutes a reliable reference point. The central issue is \textit{construct validity}, as existing work focuses narrowly on political preference or ideological alignment, while neglecting other dimensions that determine whether a response is reliable in a news context. Moreover, evaluation criteria are often derived from model outputs themselves \cite{openai2025politicalbias}, introducing circularity, or borrowed from instruments such as the Political Compass Test \cite{motoki2024human, rozado2024preferences, nature_liu}, which lack validation for this setting \cite{rottger-etal-2024-political, khetan2026politicsbenchbenchmarkingpoliticalvalues}. Consequently, what is being measured is underspecified, making results difficult to interpret or compare across evaluations.

These issues extend to \textit{measurement}. While automatic evaluation is necessary for scale, its validity depends on stable and well-justified reference labels. Current approaches frequently rely on public opinion polling \cite{OpinionQA, openai2025politicalbias} or non-expert crowd annotations \cite{wang2026theexplanationmakessense}, neither of which constitutes a validated standard for assessing the quality of news outputs. Moreover, evaluation inputs are often drawn from narrow user populations or constructed synthetically \cite{anthropic2025evenhandedness, pct, OpinionQA, openai2025politicalbias}, limiting robustness. This motivates an alternative paradigm in which expert judgment is used not only to assess model outputs, but to define the underlying standards of high-quality news responses.

\textbf{Therefore, we introduce \ourbench{}, an expert-guided benchmark for news response quality grounded in editorial standards.} Evaluating news response quality requires forms of judgment that have historically resisted automation, including reasoning under uncertainty, weighing incomplete or conflicting evidence, and assessing claims in high-stakes contexts. In fields such as journalism, intelligence analysis, and policy research, high-quality outputs are governed by editorial standards, which are shared criteria that determine how evidence is sourced, how claims are verified, and how information is presented. We adopt this paradigm and operationalize evaluation along three core dimensions: \textit{source quality}, \textit{factuality}, and \textit{neutrality}. Together, these dimensions capture whether a model relies on credible evidence, produces verifiable claims, and avoids framing that implicitly advances particular political positions.

We operationalize this paradigm through a multi-stage pipeline that translates expert judgment into standardized and scalable evaluation. We begin by conducting structured interviews with senior experts to define the core dimensions of news response quality, establishing the foundations of an editorial standard. We then work with mid-tier experts to iteratively develop and calibrate a set of editorial guidelines that formalize these standards into concrete, operational procedures, ensuring that independent experts can apply them with high agreement. These guidelines serve as the reference policy for evaluation, which we subsequently use to align and calibrate automatic AI judges to replicate expert application of the standards at scale. In parallel, experts guide the construction of evaluation inputs through a hybrid approach that combines real-world prompts, targeted synthetic coverage, and expert-identified edge cases. Finally, \ourbench{} incorporates an adaptive component, with a subset of inputs continuously updated to reflect current events, allowing the evaluation to remain aligned with the evolving information environment. Together, this pipeline reframes evaluation as the enforcement of expert-defined standards, producing a benchmark that is more grounded, reliable, and responsive than prior work.

We show that this pipeline yields editorial standards that are both highly reliable across expert annotators, as evidenced by strong inter-rater agreement, and readily transferable to automatic AI judges. The resulting \ourbench{} benchmark comprises three key artifacts:
\begin{enumerate}
\item A test dataset of ${\sim}3K$ inputs spanning real-world prompts, expert-generated cases, and targeted synthetic examples on political topics, with ${\sim}500$ refreshed monthly to reflect current events.
\item A gold-labeled dataset of ${\sim}2.5K$ expert-annotated evaluation items spanning prompt--response, claim-level, and source-level judgments.
\item A suite of automatic AI judges calibrated and validated against expert annotations.
\end{enumerate}
Together, these components provide a scalable, expert-grounded benchmark for evaluating the quality of AI-generated news responses.

%% file: sections/2_method.tex
\section{\ourbench{} Overview}

\ourbench{} evaluates the quality of AI-generated political news responses using a pipeline grounded in expert-defined standards rather than proxy signals or assumed ground truth. The benchmark is built in two phases. In Phase 1, domain experts define editorial standards and produce gold-label datasets aligned with those standards. In Phase 2, these standards and labels are used to calibrate automated AI evaluators that can apply expert judgment at scale.

\subsection{Evaluation Dimensions and Prompt Construction}\label{sect:prompt_construction}

High-quality political news responses are inherently multi-dimensional. Rather than reducing quality to a single construct such as political bias, \ourbench{} focuses on three dimensions that experts consistently identified as central to trustworthy information: \textbf{source quality}, \textbf{factuality}, and \textbf{neutrality}. Source quality evaluates whether evidence is epistemically reliable and appropriate, factuality assesses whether claims are accurate and verifiable, and neutrality measures whether information is presented without politically skewed framing. These dimensions are operationalized through explicit editorial standards developed by experts.

To evaluate models across realistic and challenging settings, \ourbench{} combines three complementary prompt sources. \textit{Real-world prompts} are drawn from online spaces where users discuss political topics (e.g., Reddit, social media, forums) and retain natural phrasing. \textit{Expert-generated prompts} are designed to probe rare but important failure modes, including adversarial framing and politically sensitive edge cases. \textit{Synthetic prompts} are used to improve coverage, balance politically asymmetric prompts, create adversarial variants, and ensure timely evaluation of emerging political events. Together, these sources yield a dataset that is both representative of real-world usage and systematically designed to surface model weaknesses.

Lastly, since political events evolve rapidly, the benchmark combines an ``evergreen'' component that remains stable across evaluation cycles with a rotating ``current-events'' component refreshed regularly to reflect emerging political developments.

\Cref{tab:input_breakdown,tab:final_data_breakdown} show the breakdown of number of inputs from each source and adaptive component.
\input{table/input_breakdown}
\input{table/final_dataset_breakdown}

\subsection{Phase 1: Expert-Guided Standards and Gold Labels}\label{sect:phase1}

In Phase 1, we recruit domain experts and conduct extensive structured interviews to distill the tacit evaluative skills used in real-world political information assessment. This phase produces two core artifacts: (1) editorial standards, curated by senior experts, that define how responses should be evaluated and (2) a gold-label dataset, annotated by domain experts, used to calibrate automated judges.

To develop standards, we recruit experts through a tiered structure. \textbf{Tier 1 experts} are senior practitioners with substantial experience making high-stakes political judgments, including senior journalists, policy experts, intelligence professionals, and government leaders. Their role is to define standards and adjudicate difficult cases. \textbf{Tier 2 experts} are domain specialists with deep expertise in specific subject areas who apply the standards to evaluate responses at scale. Tier 2 experts include a political economist at a leading think tank, a regional specialist at a foreign policy institute, and an investigative journalist who has covered a particular beat for decades.

Additionally, cross-perspective credibility is a structural requirement. If the expert layer used to calibrate the system is perceived as reflecting a single viewpoint or constituency, confidence in the evaluation framework breaks down. Accordingly, for each domain, we assemble expert panels spanning relevant axes of professional and intellectual diversity to ensure that evaluation criteria reflect shared analytical standards rather than the preferences of any particular faction.

The standards development process begins with structured interviews with many Tier 1 experts to elicit the principles they use when evaluating political information quality. In these sessions, we focus specifically on the implicit criteria that guide them when making high-stakes decisions. These insights are consolidated into initial editorial guidelines and iteratively refined through calibration exercises with Tier 2 experts until agreement stabilizes. Rather than asking evaluators whether a response is personally ``good'' or ``biased,'' the process emphasizes whether responses adhere to explicit standards, improving reproducibility across raters.

Using the finalized guidelines, Tier 2 experts independently annotate prompt--response pairs to create a gold-label dataset. Disagreements are reviewed through structured adjudication, with ambiguous or contested cases escalated to Tier 1 experts when necessary. This process yields expert-consistent labels while also identifying edge cases that inform targeted refinements to the guidelines.

\subsection{Phase 2: Calibration of Automated AI Judges}

While expert evaluation provides high-quality judgments, it is too costly to scale to ongoing model assessment. In Phase 2, we therefore calibrate automated AI judges to apply the same standards consistently.

We begin with the editorial guidelines developed in Phase 1 and iteratively refine equivalent specifications for an AI judge using signal from the expert gold-label dataset. At each iteration, the AI judge evaluates held-out examples and its agreement with expert annotations is measured. Low-agreement areas are identified, and the AI evaluator instructions are refined, using an AI coding agent to propose targeted modifications. This process continues until performance stabilizes or predefined agreement thresholds are met.

The resulting AI judges are designed to faithfully implement expert standards while remaining as simple as the task permits. Some evaluations can be performed with a single model call, while more complex tasks require structured, multi-step reasoning. Across all tasks, the objective is consistent application of expert-defined standards at scale.

%% file: table/input_breakdown.tex
\begin{table}[h]
\centering
\small
\begin{tabular}{lrr}
\toprule
\textbf{Source} & \textbf{Total Number} & \textbf{Share} \\
\midrule
Real-World & 1{,}500 & 47.8\% \\
Expert-Generated & 750 & 23.9\% \\
Synthetic & 885 & 28.2\% \\
\midrule
\textbf{Total} & \textbf{3{,}135} & \textbf{100.0\%} \\
\bottomrule
\end{tabular}
\caption{Breakdown of NewsBench input sources in the final prompt set.}
\label{tab:input_breakdown}
\end{table}

%% file: table/final_dataset_breakdown.tex
\begin{table}[h]
\centering
\small
\begin{tabular}{lrrrrr}
\toprule
\textbf{Component} & \textbf{Real-World} & \textbf{Expert-Generated} & \textbf{Synthetic} & \textbf{Total} & \textbf{Share} \\
\midrule
Evergreen & 966 & 664 & 433 & 2{,}063 & 65.8\% \\
Current Events & 534 & 86 & 452 & 1{,}072 & 34.2\% \\
\midrule
\textbf{Total} & \textbf{1{,}500} & \textbf{750} & \textbf{885} & \textbf{3{,}135} & \textbf{100.0\%} \\
\bottomrule
\end{tabular}
\caption{Breakdown of evergreen and current events components in the final NewsBench prompt set, by source type.}
\label{tab:final_data_breakdown}
\end{table}

%% file: sections/3_results_validation.tex
\section{Validation of Methodology}
In this section, we evaluate the reliability of the expert-defined editorial guidelines developed in Phase 1 and assess whether the calibrated AI judges from Phase 2 serve as suitable proxies for expert evaluation across all three domain tasks: source quality, factuality, and neutrality. For each task, we construct validation and held-out test datasets using the prompts described in \cref{sect:prompt_construction}, paired with responses sampled across major frontier model families, including OpenAI GPT, Anthropic Claude, Google Gemini, and xAI Grok. Some task-specific validation data also includes search-augmented systems. This model diversity is intentionally used to evaluate whether the AI judges remain aligned with expert annotation across outputs produced by different systems.

\paragraph{Metrics} For each task, we first measure human--human reliability using Krippendorff's $\alpha$, which assesses whether the editorial guidelines can be applied consistently across independent expert evaluators. We then evaluate AI judges on a held-out test set to avoid optimistic bias from calibration. This test set is constructed during Phase 1 using annotations from at least two Tier 2 experts; when disagreements arise, a Tier 1 expert adjudicates the final gold label. Finally, we report precision, recall, and F1 of AI judge evaluations relative to these expert-adjudicated gold labels.

\subsection{Source Quality}
The Source Quality task evaluates the credibility of sources cited in an AI response. We first use a foundational AI model to extract cited sources from each generation using prompting. Then, each extracted source is evaluated using the two core standards identified in Phase 1:

\begin{itemize}
        \item \textit{Source Type}: A six-level standard identifying the type of cited source (``primary'', ``scholarly'', ``think tank'', ``journalistic'', ``commercial content'', ``informal'').
        \item \textit{State-Controlled Flag}: A binary standard indicating whether a cited outlet is a media organization whose editorial independence is compromised by state control (e.g., RT, CGTN, Xinhua, Al Jazeera). This determination is made independently of source type. Government websites publishing official records are not flagged, nor are editorially independent state-funded broadcasters (e.g., BBC, NPR, PBS, Deutsche Welle).
\end{itemize}
We then convert these structured judgments into a deterministic source-quality score on a $0-100$ scale using an algorithm developed from expert guidance. Responses with no cited sources are excluded from source-quality scoring rather than assigned a score of zero. In the current implementation, we use only source type and the State-Controlled flag in the final score. Additional deductions for fabricated sources and sources outside their domain of competence are planned for future versions.

\paragraph{AI Judge Calibration}
For Phase 2 AI judge calibration, we used the gold-label validation dataset produced in Phase 1 ($n \approx 230$). The base source-quality AI judge used was GPT-5.2 \cite{gpt52}.

\paragraph{Test Dataset} For final evaluation, we used a new held-out test set ($n \approx 150$). Gold labels were obtained using Tier 2 experts who did not participate in Phase 1 development. Each expert independently evaluated their assigned citations using the final editorial guidelines from Phase 1. When all evaluators agreed, their shared judgment was retained, but in cases of disagreement, a Tier 1 expert adjudicated the final outcome.

\subsubsection{Results: Source Quality}
\paragraph{Editorial guidelines reliably generalize across experts.}
\Cref{tab:sourcequality_human_human} shows that, across both standards, the independent expert evaluators achieve high inter-rater reliability, with $\alpha = 0.83$ on Source and $\alpha = 0.69$ on State-Controlled. We note that the inter-rater agreement for the Source standard on the validation set changed from $\alpha = 0.74$ under the initial guidelines to $\alpha = 0.82$ under the refined guidelines produced in Phase 1, indicating improved consistency following iterative rubric development (described in \cref{sect:phase1}). This suggests that the Phase 1 pipeline was important for producing editorial guidelines that support reliable annotation. More broadly, the observed level of agreement indicates that the rubric generalizes beyond its original authors and can be consistently applied by independent human experts.

\input{table/results_sourcequality_human_human}

\input{table/results_sourcequality_humanAI_PRF1}

\paragraph{AI judges closely match expert application of standards.} 
\Cref{tab:sourcequality_human_AI_PRF1} shows that, across both standards, the AI judge attains consistently high F1 scores (macro F1 = 0.75 on Source and 0.95 on State-Controlled) relative to expert annotations, including near-perfect performance on the State-Controlled standard (macro F1 = $0.95$; weighted F1 = $0.97$). Overall, these results indicate that the calibrated AI judge successfully reproduces expert application of the editorial standards, enabling reliable and consistent evaluation at scale. These findings support the use of AI judges for scalable Source Quality assessment.

\subsection{Factuality}
The Factuality task evaluates whether the claims in an AI-generated response are grounded in verifiable information. Following prior work such as FActScore \cite{min-etal-2023-factscore}, SAFE \cite{wei2024longformfactuality}, and VeriScore \cite{song-etal-2024-veriscore}, and after validation with Tier 1 experts, we decompose this task into two steps. First, we extract verifiable claims from a response and then independently assess the validity of each claim.

To extract claims, we employ an over-extraction and filtering pipeline, as is common in similar tasks. We first use an AI model to identify all \textit{potentially} verifiable claims within each response. To ensure extraction quality, each candidate claim is filtered according to four expert-validated criteria: claims must be standalone (non-self-referential), sufficiently specific to permit independent verification, non-compound (containing a single independently verifiable assertion), and faithful to the source response (preserving meaning without omission or addition). This extractor is then applied to all responses, and filtered claims are retained for downstream evaluation.

We next evaluate claim verification, which assesses whether an extracted claim is supported by verifiable evidence. Because this task is operationally closer to journalistic fact-checking than open-ended editorial judgment, Tier 2 domain specialists with fact-checking experience led standard development and annotation. Through Phase 1, experts established a binary verification standard in which raters assess whether a claim, given the original prompt, source response, and generation date, is supported by available evidence. Providing the response generation date ensures that time-sensitive claims are evaluated relative to the information available when the response was produced.  Raters verify claims against external sources, reserving ``false'' for clear, evidence-backed contradictions and otherwise assigning ``true,'' including in cases of ambiguity or legitimate disagreement.

\paragraph{AI Judge Calibration}
In Phase 2, we calibrated the claim verification judge to reproduce expert judgment using the Phase 1 standards and the gold-labelled validation set ($n\approx 200$). The claim verification judge is an agent that uses a combination of GPT-5.2 \cite{gpt52} and GPT-5.5 \cite{gpt55}. It evaluates each claim in the context of the original prompt, the full model response, and the response generation date. When needed, the AI judge may consult external evidence from the internet before determining whether the claim is supported or unsupported, similar to human experts.

\paragraph{Test Dataset}
For evaluation, we use claim-level validation from a held-out test set labeled by Tier 2 fact-checking experts ($n \approx 300$). As before, any non-consensus labels are adjudicated by Tier 1 experts. In the full dataset, supported claims substantially outnumber unsupported claims. To ensure robust evaluation across both verdict types, we construct the held-out test set using stratified sampling to include 20\% unsupported claims, enabling more reliable assessment of performance on both supported and unsupported judgments.

\subsubsection{Results: Factuality}

\paragraph{Claim factuality is a hard task even for domain experts.}
\Cref{tab:factuality_human_human} shows that expert evaluators achieve only moderate agreement ($\alpha = 0.39$ on test set) on the claim verification task. However, importantly, because our annotation process captures not only final labels but also expert reasoning, we are able to diagnose the source of disagreement. This reveals a systematic pattern in which \textit{most divergence arises not from differences in factual assessment itself, but from thresholding decisions about how strictly to apply the standard}. For example, given the claim “Most states openly say the UN Security Council’s structure has a serious legitimacy deficit as of May 11, 2026,” experts may reasonably disagree on whether “most states” is sufficiently supported. \textbf{Given this relatively low inter-rater reliability, we rely on Tier 1 experts to adjudicate gold labels and provide additional guidance to ensure consistent application of standards for ambiguous claims.}

\input{table/results_factuality_human_human}
\input{table/results_factuality_humanAI_PRF1}

\paragraph{\textbf{AI judges achieve moderate agreement with experts.}}
\Cref{tab:factuality_human_AI_PRF1} shows that AI judges achieve moderate alignment with expert annotations on the claim verification task (macro $F1 = 0.69$, weighted $F1 = 0.75$). Examining performance by label, however, reveals that the AI judge is substantially more reliable when confirming supported claims than when assigning falsity, suggesting a tendency toward over-rejection in which some claims judged valid by experts are incorrectly marked as unsupported.

In a Tier 1 fact-checker review of over $200$ cases where the AI judge flagged claims as false but experts marked them true, we found that $60\%$ were either correctly flagged or involved meaningful ambiguity with identifiable technical or evidentiary issues, such as saying ``all restrictions'' on Iran’s missile program had expired when some remained. When combined with the accurately flagged false claims, this means that \textbf{roughly $85\%$ of false flags reflected correct or meaningfully defensible judgments}.

\subsection{Neutrality}
The Neutrality task evaluates whether a model response favors a specific partisan position. At Forum, we recognize that ``neutrality'' is inherently context-dependent and does not admit a single scalar definition. Therefore, rather than attempting to capture it directly, we used the structured interviews from Tier 1 experts in Phase 1 to develop a set of editorial standards that jointly serve as reliable indicators of neutrality. Through this process, experts converged on a flowchart-based decision procedure that first classifies prompts by type and then routes them to type-specific standards, reflecting the insight that neutrality is a property of the prompt-response pair rather than the response alone. The flowchart terminates in a binary verdict of ``approximate neutrality,'' reflecting the fact that no rule set fully resolves the underlying conceptual ambiguity. 

For responses that do not meet the initial Neutrality standards, we conduct a secondary analysis to characterize the direction of any partisan lean. At present, this evaluation is limited to U.S. political orientations, specifically left or right, though we plan to extend it to additional dimensions such as globalist versus anti-globalist or Western versus non-Western perspectives. Consistent with the prior stage, this process begins by identifying the response type and then applying a corresponding flowchart of expert-informed standards. This procedure yields a three-way classification: left-leaning, right-leaning, or no clear lean, where the latter captures mixed signals, unmodeled dimensions of bias, or cases that are too subtle to assess confidently.

Although the full flowcharts are proprietary, we describe the prompt categories and a subset of the more subjective editorial standards developed for transparency.

Our evaluation framework first classifies prompts into categories (e.g., factual, directed, loaded, normative, and open-ended), each with distinct standards reflecting that neutral behavior depends on prompt type. Neutrality is operationalized through a set of editorial standards developed by Tier 1 experts and grounded in norms from journalism, government, and political institutions, with named rules, concrete examples, and reproducibility tests. Core principles include faithfully executing directed prompts, correcting loaded premises without overcorrection, balancing normative questions, and avoiding one-sided framing disguised as nuance. For responses that fail neutrality, Political Lean is assessed using a parallel rule-based framework that attributes directional lean based on factors such as accepted premises, overcorrections, framing choices, and selective refusals, while accounting for subtler signals like omission and asymmetric burden of proof. 

We recognize that these rules are themselves debatable; reasonable people can disagree about where the lines fall. We make no claim that these are the only or best definitions of neutrality. Rather, they were developed by domain experts, and we document them here in the interest of transparency about how our evaluation operates.

\paragraph{AI Judge Calibration} The Phase 1 gold set includes approximately 300 prompt–response pairs for Neutrality and 200 for Political Lean, each with gold-label verdicts. The selected Neutrality evaluator is a DSPy \cite{khattab2024dspy} monolithic program run with Claude Opus 4.6 \cite{anthropic_claude_opus_4_6}; the selected Political Lean evaluator is a DSPy decision-tree program run with GPT-5.4 \cite{gpt5_4}. We select evaluator models independently by task based on held-out validation performance and implementation fit, so the base model differs from the GPT-5.2 source-quality judge.

\paragraph{Test Dataset}
For evaluation, we used two new held-out test sets with approximately $n\approx 160$ pairs for Neutrality and $n\approx 100$ pairs for Political Lean. Tier 2 experts independently produced gold labels, with disagreements resolved by a Tier 1 expert acting as final adjudicator.

\subsubsection{Results: Neutrality}
\paragraph{Editorial standards constructed in Phase 1 yield consistent expert agreement.}
\Cref{tab:neutrality_human_human} shows that expert evaluators achieve high inter-rater reliability, with a test set $\alpha = 0.72$ and validation set $\alpha = 0.79$. These results indicate that the Phase 1 pipeline yields a verifiable standard that is applied consistently across independent experts, and produces a validation set that is stable for calibration in Phase 2.

\input{table/results_neutrality_human_human}
\input{table/results_neutrality_humanAI_PRF1}

\paragraph{The AI judge achieves substantial alignment with human neutrality verdicts.} \Cref{tab:neutrality_human_AI_PRF1} shows strong overall performance, with a weighted F1 of $0.86$ and a macro F1 of $0.85$. The AI judge also achieves high precision and recall, indicating strong performance across both positive and negative classifications. Overall, these results suggest that the AI judge closely tracks human expert neutrality judgments.

\input{table/results_lean_human_human}
\input{table/results_lean_humanAI_PRF1}

\paragraph{Political Lean judgments are reproducible and the AI judge captures the main directional signal.}
\Cref{tab:lean_human_human} shows that expert evaluators apply the Political Lean framework with moderate agreement, with $\alpha = 0.68$ on the validation set and $\alpha = 0.65$ on the held-out test set. This is lower than the binary Neutrality task, which is expected because the Political Lean task requires assigning direction among three classes after identifying a neutrality failure.

\Cref{tab:lean_human_AI_PRF1} reports a weighted F1 of $0.79$ and a macro F1 of $0.73$ for the calibrated AI judge against adjudicated human expert labels on the held-out test set. Overall, this indicates that the calibrated AI judge can reliably mimic the annotations of the human experts.

%% file: table/results_sourcequality_human_human.tex
\begin{table}[h]
\centering
\small
\begin{tabular}{lcc}
\toprule
& \multicolumn{1}{c}{\textbf{Validation Set}} & \multicolumn{1}{c}{\textbf{Test Set}}\\
\cmidrule(lr){2-2} \cmidrule(lr){3-3}
\textbf{Standard} & $\alpha$ [95\% CI] & $\alpha$ [95\% CI] \\
\midrule
Source (\textit{categorical}) & .82 [.76, .87] & .83 [.76, .89] \\
State-Controlled (\textit{binary}) & .93 [.80, 1.00] & .69 [.52, .83] \\
\bottomrule
\end{tabular}
\caption{Human--human inter-rater reliability for the Source Quality standards, reported per set. The validation pairs were labeled during Phase 1 calibration; the test set was held out from both Phase 1 and 2. We report human-human reliability via Krippendorff's $\alpha$, and 95\% confidence intervals computed via item-bootstrap with $B=2,000$.}
\label{tab:sourcequality_human_human}
\end{table}

%% file: table/results_sourcequality_humanAI_PRF1.tex
\begin{table}[h]
\centering
\small
\begin{tabular}{lccc}
\toprule
\textbf{Standard} & \textbf{Precision} & \textbf{Recall} & \textbf{F1} \\
\midrule
\multicolumn{4}{l}{Source (\textit{categorical}) }\\
\quad \textit{macro avg} & .74 & .80 & .75 \\
\quad \textit{weighted avg} & .86 & .79 & .81 \\
\midrule
\multicolumn{4}{l}{State-Controlled (\textit{binary}) } \\
\quad \textit{macro avg} & .98 & .92 & .95 \\
\quad \textit{weighted avg} & .97 & .97 & .97 \\
\bottomrule
\end{tabular}
\caption{Performance of the AI judge on Source Quality standards using the adjudicated human labels as gold labels. We report precision, recall, and F1 using a \textit{macro avg} and \textit{weighted avg}. For Source, \textit{macro avg} is the unweighted mean across the six tiers; \textit{weighted avg} is support-weighted.}
\label{tab:sourcequality_human_AI_PRF1}
\end{table}

%% file: table/results_factuality_human_human.tex
\begin{table}[h]
\centering
\small
\begin{tabular}{lcc}
\toprule
& \multicolumn{1}{c}{\textbf{Validation Set}} & \multicolumn{1}{c}{\textbf{Test Set}}\\
\cmidrule(lr){2-2} \cmidrule(lr){3-3}
\textbf{Standard} & $\alpha$ [95\% CI] & $\alpha$ [95\% CI] \\
\midrule
Claim Verification (\textit{binary}) & .47 [.36, .58] & .39 [.26, .51] \\
\bottomrule
\end{tabular}
\caption{Human--human inter-rater reliability for Factuality, reported per set. The validation samples were labeled during Phase 1 calibration; the test set was held out from both Phase 1 and 2. We report human-human reliability via Krippendorff's $\alpha$, and 95\% confidence intervals computed via item-bootstrap with $B=2,000$.}
\label{tab:factuality_human_human}
\end{table}

%% file: table/results_factuality_humanAI_PRF1.tex
\begin{table}[h]
\centering
\small
\begin{tabular}{lccc}
\toprule
\textbf{Standard} & \textbf{Precision} & \textbf{Recall} & \textbf{F1} \\
\midrule
\multicolumn{4}{l}{Claim Verification (\textit{binary})}\\
\quad \textit{macro avg} & .69 & .75 & .69 \\
\quad \textit{weighted avg} & .80 & .74 & .75 \\
\bottomrule
\end{tabular}
\caption{Performance of the AI judge on Factuality using the adjudicated human labels as gold labels. We report precision, recall, and F1 using a \textit{macro avg} and \textit{weighted avg}.}
\label{tab:factuality_human_AI_PRF1}
\end{table}

%% file: table/results_neutrality_human_human.tex
\begin{table}[h]
\centering
\small
\begin{tabular}{lcc}
\toprule
& \multicolumn{1}{c}{\textbf{Validation Set}} & \multicolumn{1}{c}{\textbf{Test Set}}\\
\cmidrule(lr){2-2} \cmidrule(lr){3-3}
\textbf{Standard} & $\alpha$ [95\% CI] & $\alpha$ [95\% CI] \\
\midrule
Neutrality (\textit{binary}) & .79 [.66, .90] & .72 [.60, .83] \\
\bottomrule
\end{tabular}
\caption{Human--human inter-rater reliability for Neutrality, reported per set. The validation pairs were labeled during Phase 1 calibration; the test set was held out from both Phase 1 and 2. We report human-human reliability via Krippendorff's $\alpha$, and 95\% confidence intervals computed via item-bootstrap with $B=2,000$.}
\label{tab:neutrality_human_human}
\end{table}

%% file: table/results_neutrality_humanAI_PRF1.tex
\begin{table}[h]
\centering
\small
\begin{tabular}{lccc}
\toprule
\textbf{Standard / Class} & \textbf{Precision} & \textbf{Recall} & \textbf{F1} \\
\midrule
\multicolumn{4}{l}{Neutrality (\textit{binary})}\\
\quad \emph{macro avg} & .85 & .85 & .85 \\
\quad \emph{weighted avg} & .86 & .86 & .86 \\
\bottomrule
\end{tabular}
\caption{Performance of the AI judge on the Neutrality standard using the adjudicated human labels as gold, evaluated on the test set. We report precision, recall, and F1 using a \emph{macro avg} and \emph{weighted avg}.}
\label{tab:neutrality_human_AI_PRF1}
\end{table}

%% file: table/results_lean_human_human.tex
\begin{table}[h]
\centering
\small
\begin{tabular}{lcc}
\toprule
& \multicolumn{1}{c}{\textbf{Validation Set}} & \multicolumn{1}{c}{\textbf{Test Set}}\\
\cmidrule(lr){2-2} \cmidrule(lr){3-3}
\textbf{Standard} & $\alpha$ [95\% CI] & $\alpha$ [95\% CI] \\
\midrule
Political Lean (\textit{categorical}) & .68 [.55, .80] & .65 [.51, .78] \\
\bottomrule
\end{tabular}
\caption{Human--human inter-rater reliability for Political Lean, reported per set. The validation samples were labeled during Phase 1 calibration; the test set was held out from both Phase 1 and 2. We report human-human reliability via Krippendorff's $\alpha$, and 95\% confidence intervals computed via item-bootstrap with $B=2,000$.}
\label{tab:lean_human_human}
\end{table}

%% file: table/results_lean_humanAI_PRF1.tex
\begin{table}[ht!]
\centering
\small
\begin{tabular}{lccc}
\toprule
\textbf{Standard / Class} & \textbf{Precision} & \textbf{Recall} & \textbf{F1} \\
\midrule
\multicolumn{4}{l}{Political Lean (\textit{categorical}) }\\
\quad \emph{macro avg} & .72 & .76 & .73 \\
\quad \emph{weighted avg} & .81 & .78 & .79 \\
\bottomrule
\end{tabular}
\caption{Performance of the AI judge on the Political Lean standard using the adjudicated human labels as gold, evaluated on the test set. We report precision, recall, and F1 using a \emph{macro avg} and \emph{weighted avg}.}
\label{tab:lean_human_AI_PRF1}
\end{table}

%% file: sections/4_results_AIpreformance.tex
\section{Current AI Performance on \ourbench{}}
In this section we use \ourbench{} to evaluate the behavior of four foundation AI models. The evaluation set covers 3{,}135 prompts answered by ChatGPT (GPT-5.5), Gemini 3.1 Pro, Claude Opus 4.7, and Grok 4.3, producing ${\sim}12{,}500$ model responses. Each response is evaluated on three dimensions: whether it remains neutral on contested topics, whether its verifiable claims are factually supported, and whether its cited sources are credible. These results reflect model versions as of May 10, 2026.

\paragraph{Response generation settings.}
Responses were generated through each provider's API using a shared response-generation harness. All four models were run with the same system instruction, which directed the model to answer accurately and concisely and to say when it was uncertain, and with web search made available for current or factual information. The production run used our \texttt{high} generation configuration for each model: GPT-5.5 was called as \texttt{gpt-5.5-2026-04-23} with high reasoning effort; Claude Opus 4.7 was called as \texttt{claude-opus-4-7} with adaptive thinking at high effort and \texttt{max\_tokens=128000}; Gemini 3.1 Pro was called as \texttt{gemini-3.1-pro-preview} with \texttt{thinkingLevel=HIGH}; and Grok 4.3 was called with high reasoning effort. No non-default temperature was specified for these \texttt{high} configuration runs. The harness used a 300-second idle timeout and a 15-minute wall-clock ceiling per request.

\paragraph{No model dominates across all dimensions.}
\Cref{tab:model_results_overview} shows that models exhibit distinct failure profiles. Gemini is the most balanced performer across the three axes and is the only model that ranks in the top two for neutrality, factual accuracy, and source quality. In contrast, Grok performs weakest in all tasks, with particularly low factual accuracy (only $57.1\%$ of responses pass factuality). ChatGPT and Claude occupy a middle ground with complementary strengths as ChatGPT performs best on factual accuracy ($91.1\%$ of responses pass factuality) but scores lower on neutrality and source quality, while Claude achieves the highest average source quality ($81.5$) despite lower factual accuracy.

 \paragraph{Sourcing quality and factual accuracy are not aligned.} Interestingly, we find that models with high average source-quality scores do not always have high factual scores. For example, Claude's average source-quality score is highest (81.5), yet only 58.9\% of its responses are free of false claims. Conversely, ChatGPT has the best factuality score (91.1\%) despite a lower average source-quality score (74.3) than Claude and Gemini. This pattern suggests that source presentation and factual reliability should be evaluated as separate dimensions rather than collapsed into a single quality score.

\input{table/results_model_overview}

\paragraph{Factuality separates the models sharply.}
\Cref{tab:model_results_factuality} reports more analysis of factuality by model. We find that ChatGPT is most factual with the lowest percentage ($8.9\%$) of responses with at least one false claim and the lowest false-claim rate ($0.85\%$). Gemini is second best by the response-level metric, with $24.6\%$ of responses containing at least one false claim, while Claude and Grok each have a little over $40\%$ of responses containing at least one false claim.

\input{table/results_model_factuality}

%% file: table/results_model_overview.tex
\begin{table}[htbp]
\centering
\small
\resizebox{\textwidth}{!}{%
\begin{tabular}{lrrrr}
\toprule
Model & Neutrality Pass & Factual Pass & Avg Source-Quality Score & Directional Lean Ratio \\
\midrule
Claude Opus 4.7 & 82.2\% & 58.9\% & 81.5 & 12.7x left \\
Gemini 3.1 Pro & 83.0\% & 75.4\% & 78.8 & 4.9x left \\
ChatGPT (GPT-5.5) & 74.8\% & 91.1\% & 74.3 & 15.5x left \\
Grok 4.3 & 70.1\% & 57.1\% & 73.7 & 4.2x right \\
\bottomrule
\end{tabular}
}
\caption{Model results on NewsBench. \textit{Neutrality Pass} is the percentage of responses passing the neutrality judge; \textit{Factual Pass} is the percentage of responses with no false verifiable claims; \textit{Avg Source-Quality Score} is the mean 0--100 source-quality score (higher better); \textit{Directional Lean Ratio} is the larger left/right direction among directional neutrality failures, excluding none/other cases.}
\label{tab:model_results_overview}
\end{table}

%% file: table/results_model_factuality.tex
\begin{table}[htbp]
\centering
\small
\resizebox{\textwidth}{!}{%
\begin{tabular}{lrrr}
\toprule
Model & Responses with False Claim & \# Claims / Response & False-Claim Rate \\
\midrule
Claude Opus 4.7 & 41.1\% & 17.03 & 4.30\% \\
Gemini 3.1 Pro & 24.6\% & 11.47 & 3.07\% \\
ChatGPT (GPT-5.5) & 8.9\% & 13.86 & 0.85\% \\
Grok 4.3 & 42.9\% & 19.00 & 3.62\% \\
\bottomrule
\end{tabular}
}
\caption{Factuality by model. \textit{Responses with False Claim} is the percentage of responses containing at least one false verifiable claim; \textit{\# Claims / Response} is the mean number of extracted verifiable claims per response; \textit{False-Claim Rate} is the average response-level percentage of extracted claims labeled false.}
\label{tab:model_results_factuality}
\end{table}